\documentclass[preprint,12pt]{elsarticle}

\usepackage{amssymb}
\usepackage{subfig,graphicx}
\usepackage{setspace} 
\usepackage{lineno}
\usepackage[T1]{fontenc}
\usepackage{multirow}
\usepackage{adjustbox}
\usepackage{float}
\usepackage[margin=1in]{geometry}

\journal{arxiv}

\begin{document}

\begin{frontmatter}


\title{Reconstruction and analysis of negatively buoyant jets with interpretable machine learning}

\author[inst1]{Marta Alvir}
\ead{malvir@riteh.hr}
\author[inst1,inst2]{Luka Grb\v{c}i\'c}
\ead{lgrbcic@riteh.hr}
\author[inst2]{Ante Sikirica}
\ead{ante.sikirica@uniri.hr}
\author[inst1]{Lado Kranj\v{c}evi\'c\corref{cor1}}
\cortext[cor1]{Corresponding author}

\ead{lado.kranjcevic@riteh.hr}
\affiliation[inst1]{organization={Department of Fluid Mechanics and Computational Engineering, Faculty of Engineering, University of Rijeka},
            addressline={Vukovarska 58}, 
            city={Rijeka},
            postcode={51000}, 
            country={Croatia}}

\affiliation[inst2]{organization={Center for Advanced Computing and Modelling, University of Rijeka},
            addressline={Radmile Matej\v{c}i\'c 2}, 
            city={Rijeka},
            postcode={51000}, 
            country={Croatia}}

\begin{abstract}
In this paper, negatively inclined buoyant jets, which appear during the discharge of wastewater from processes such as desalination, are observed. To minimize harmful effects and assess environmental impact, a detailed numerical investigation is necessary. The selection of appropriate geometry and working conditions for minimizing such effects often requires numerous experiments and numerical simulations. For this reason, the application of machine learning models is proposed. Several models including Support Vector Regression, Artificial Neural Networks, Random Forests, XGBoost, CatBoost and LightGBM were trained. The dataset was built with numerous OpenFOAM simulations, which were validated by experimental data from previous research. The best prediction was obtained by Artificial Neural Network with an average of R$^2$ 0.98 and RMSE 0.28. In order to understand the working of the machine learning model and the influence of all parameters on the geometrical characteristics of inclined buoyant jets, the SHAP feature interpretation method was used.

\end{abstract}


\end{frontmatter}
\newpage
\section{Introduction}

A negatively buoyant jet appears during the discharge of heavier effluent in ambient water bodies with smaller densities.
Such phenomena mostly occur during the disposal of wastewater from desalination plants, power plants, industrial factories, and cooling water discharge from liquefied natural gas (LNG) plants.

Nowadays, numerous arid and semiarid coastal regions are encountering freshwater scarcity due to population growth and deficiency of portable water, hence the number of desalination plants is significantly increased.
During the desalination process, using brackish or seawater, drinkable water is obtained, while the by-product is high-salinity concentrated effluents, so-called desalination brine, which is discharged back into the coastal waters using submerged outfalls.
Besides elevated salt concentrations, desalination brine contains traces of chemicals, such as antiscalants, flocculants and coagulants (\cite{panagopoulos2020environmental}, which can lead to environmental degradation (\cite{alameddine2007brine})).
To minimize harmful environmental effects and maximize dilution, the brine is predominantly discharged from diffusers directed upwards at an angle (\cite{de2016bioindicators}), producing negatively inclined buoyant jets.

The design of the discharge systems and condition parameters are based on detailed experimental, mathematical, and numerical investigation in order to determine the characteristics of the jet. 

Experimental investigation of inclined buoyant jets was done by numerous researchers: \cite{cipollina2005bench}, \cite{kikkert2007inclined}, \cite{ferrari2010mixing}, \cite{papakonstantis2011inclined}, \cite{papakonstantis2011inclined2}, \cite{abessi2015effect}, \cite{ardalan2018hydrodynamic}.

Mathematical models such as VISJET (\cite{cheung2000visjet}) and CORJET (\cite{doneker2001cormix}) were implemented and compared with experimental data (\cite{jirka2008improved}, \cite{lai2012mixing},  \cite{palomar2012near}). \cite{oliver2013predicting} and \cite{nikiforakis2015modified} developed a modified integral models. \cite{roberts1997mixing} obtained experimental coefficients to calculate the geometrical characteristics of dense jets. \cite{papanicolaou2008entrainment} presented top-hat and Gaussian integral models. \cite{yannopoulos2012escaping} and \cite{bloutsos2020revisiting} used Gaussian model escaping mass approach (EMA). \cite{papakonstantis2020simplified} proposed an analytical model for predicting centerlines, but all models performed similarly poor for a large range of inclination angles. 

Numerical simulations provide better predictions and detailed insight into the complexity of the buoyant jet problem than a mathematical model when both were compared to experimental data (\cite{oliver2008k}). Historically, commercial software ANSYS CFX was used  (\cite{vafeiadou2005numerical}, \cite{oliver2008k}) to study inclined buoyant jets, however, most of the newer studies utilized the open-source CFD toolbox OpenFOAM, more specifically the \textit{twoLiquidMixingFoam} multiphase solver (\cite{zhang2017large}, \cite{jiang2019turbulence}, \cite{kheirkhah2021inclined}, \cite{alvir2022openfoam}), solvers supplemented with additional advection-diffusion equations for concentration and saline (\cite{ardalan2019cfd}, \cite{tahmooresi2021effects}, \cite{gildeh2015numerical}), and finally, concentration transport equations (\cite{vafa2021effect}, \cite{tahmooresi2022application}). 

In order to model certain complex effects which occur during buoyant jet transport, experimental or numerical data is needed. \cite{shao2010mixing} experimentally investigated the effect of the bed proximity, called the Coanda effect which appears when jet discharge is placed close to an impermeable boundary, which results in reduced mixing. The height limit at which it occurs depends on the flow and nozzle angle. The Coanda effect was successfully investigated using numerical simulation in OpenFOAM by \cite{ramezani2021effect}. \cite{jiang2014mixing}, \cite{abessi2016dense} and \cite{kheirkhah2021inclined} experimentally and numerically investigated the behaviour of inclined buoyant jets in shallow waters.

Lately, machine learning has been proposed as an additional method for understanding and predicting coastal and open channel flow, including hydraulic jumps (\cite{naseri2012determination}), friction factor (\cite{yuhong2009application}), outflow over rectangular side weirs (\cite{bilhan2010application}), flows through backwater bridge constrictions (\cite{pinar2010artificial}), stability number of rubble-mound breakwaters (\cite{gedik2018least}). Also, numerous researchers applied machine learning techniques for predicting the performance, maintenance and control of desalination systems (\cite{behnam2022review}).

More commonly, numerical models previously validated with experimental data, are used to build ML datasets, because it is faster and cheaper way than an experimental approach, and it is possible to investigate a large number of different configurations. Several authors applied machine learning techniques on various CFD problems, including open channel junctions (\cite{sun2014artificial}), mixing in double-Tee pipe junctions (\cite{grbvcic2020efficient}), particle transport and filtration in porous media (\cite{marcato2021computational}), cavitation model parameters prediction (\cite{sikirica2020cavitation}), drag reduction of underwater serial multi-projectiles (\cite{huang2022machine}), large gas engine prechambers (\cite{posch2021development}) and bubble column reactor (\cite{pelalak2021influence}).

Most previous authors applied machine learning to coastal outfalls with a significantly reduced number of variables and small variations of parameters. \cite{abdel2007investigation} applied a neuro-fuzzy approach to predict locations of interest for multiple outfall discharges. \cite{musavi2013simulation} used Artificial Neural Networks (ANN) to predict the upper and inferior limits of the jet trajectory of negatively buoyant jets based on four dimensionless parameters. Multigene genetic-programming (MGGP) was applied for predicting dilution of vertical buoyant jets (\cite{yan2019multigene}) and multiport buoyant discharges (\cite{yan2020evolutionary}, \cite{yan2021simulations}). \cite{yan2022cfd} applied convolutional neural network (CNN) on rosette diffusers for buoyant jets. \cite{jain2022applications} used the ANFIS model coupled with Genetic Algorithm, Particle Swarm Optimization, and Firefly Algorithm for predicting four variables of inclined buoyant jets based on Froude number and angle with numerical simulations.

In this paper, a machine learning (ML) approach for negatively inclined buoyant jets reconstruction is proposed. The dataset for the ML model is generated using CFD in order to analyze the behaviour and predict 5 geometrical characteristics of jets. 
More specifically, the analysis includes 500 simulations with varied water and pipe heights, a combination of fluid densities, discharge diameters and velocities.  Most of the previous research was focused on a small number of inclination angles, most often 30$^{\circ}$, 45$^{\circ}$ and 60$^{\circ}$. In this paper, results are obtained for a large range of angles varying from 5$^{\circ}$ to 80$^{\circ}$. The CFD model is validated with experimental data from previous research.

Several ML algorithms are used to train the models including Support Vector Regression, XGBoost, CatBoost, LightGBM, Random Forests and ANN, to get full insight into the capabilities of each model. Most of the previous research used a small number of ML algorithms, so in this paper, a comparison of their performance is shown. Prediction of ML models gives a fast and comprehensible perception of the plume behaviour for specific flow conditions and can be used as input data for far field models. 

The objective of the detailed investigation and application of ML models is to provide insight into the comprehensive behaviour of buoyant jets for a large number of different flow parameters using the ML feature interpretation framework SHAP. To the best of the authors’ knowledge, the SHAP feature explanation is applied for the first time on buoyant jets, coastal flow and outfalls. 
Ultimately, an investigation of specific cases is presented including shallow water and the Coanda effect. Results of ML models and SHAP interpretation for the dataset supplemented with past experimental data are shown. Since most of the previous research used a reduced number of variables, the prediction capabilities of ML models for this type of problem are also presented.

\section{Materials and Methods}
\subsection{Problem description}

\begin{figure}[H]
\centering
\includegraphics[width=\textwidth]{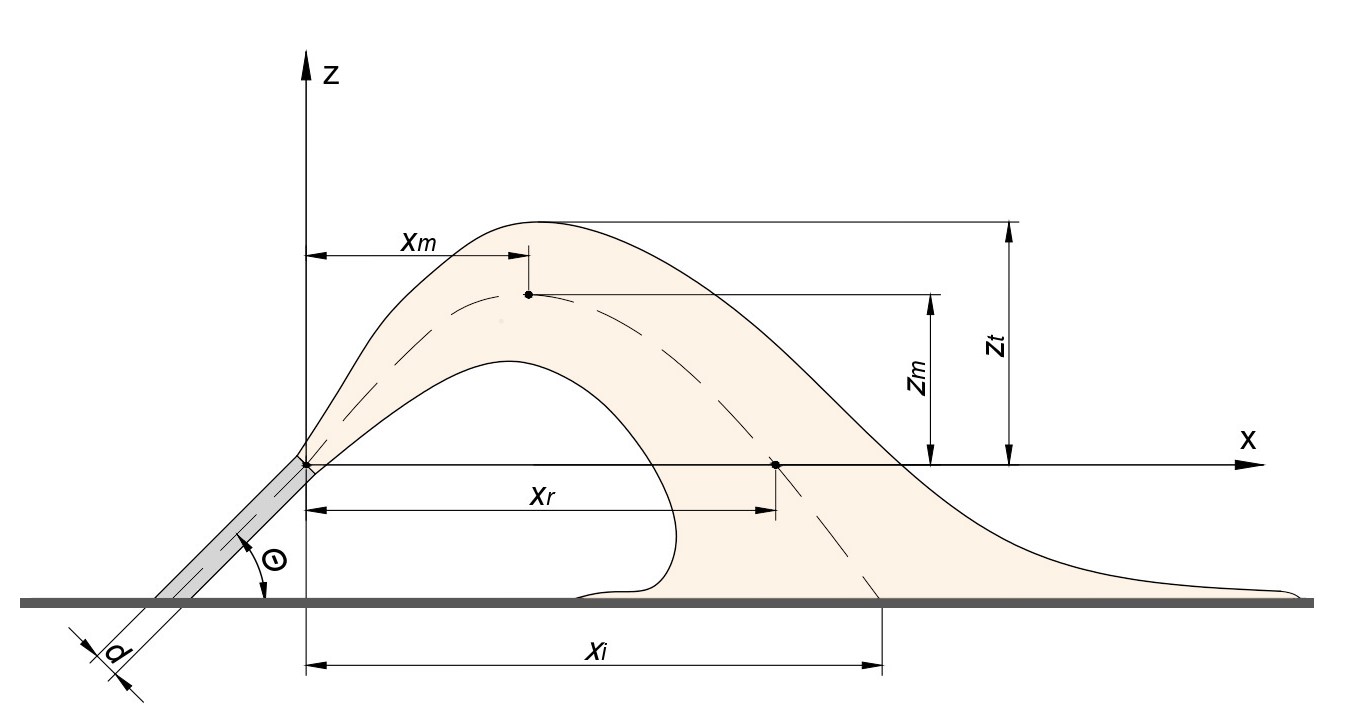}
\caption{Schematic diagram of inclined dense jets and characteristic points.}
\label{fig:points}
\end{figure}

During the discharge of negatively buoyant jets, brine density $\rho_{b}$ is bigger than ambient density $\rho_{a}$, so momentum forces act in opposite directions than gravity. The side view for the schematic diagram of negatively inclined buoyant jets is illustrated in Fig. \ref{fig:points}. The jet is released through a round nozzle inclined at angle $\theta$ with diameter $d$. Initial velocity $U_0$ is creating the initial vertical momentum flux, the jet is moving upward. After it reaches a maximum height, buoyant forces become more dominant and the jet impinges the bottom, spreading by density current (\cite{roberts1997mixing}). The centre of a coordinate system is placed in the middle of the nozzle. 

The dashed curve line presents the centerline or trajectory of buoyant jets, connecting the locus of maximum time-averaged concentration or velocities for the various cross-sections. Previous research has shown a small difference between centerlines created from velocity and contours (\cite{shao2010mixing}), so in this paper, the concentration centerline is used for further investigation, similarly to \cite{zhang2016}. Return point location $x_r$ is defined on the position where the buoyant jet passes the nozzle tap elevation, determined with ($x_r$, 0) coordinates. Horizontal and vertical locations of the centerline peak $x_m$ and $z_m$ represent the maximum location of the centerline. Terminal rise height $z_t$ is defined as a location where the concentration drops to 5 \% of the centerline peak concentration, similarly to \cite{jiang2014mixing}. Additionally, the impact point location with horizontal coordinate $x_i$ represents the point where the centerline hits the bottom.

The behaviour of negative inclined buoyant jets, assuming Boussinesq approximation ($\rho_0-\rho_a$)<<$\rho_a$, can be described with discharge angle $\theta$, 
momentum $M=U_0Q$, buoyancy $B=g_0$’$Q$, jet kinematic fluxes of volume $Q=\pi d^2U_0/4$ and acceleration due to gravity $g_0$’$=g|\rho_b-\rho_a|/\rho_a$.  Previous studies have shown a strong correlation between jet behaviour and densimetric Froude $Fr$ number which can be expressed as: 

\begin{equation}\label{eq:1}
Fr=\frac{U_0}{\sqrt{g_0'd}}
\end{equation}

Most of the previous research used the momentum-buoyancy length scale $L_M$ to compare results with different parameters of the flow. It is represented by the distance in which momentum is more significant than buoyancy on the behaviour of the jet, which can be described as follows (\cite{fischer1979mixing}): 

\begin{equation}\label{eq:2}
L_M=\frac{M^{3/4}}{B^{1/2}}=\left(\frac{\pi}{4}\right)^{1/4}dFr
\end{equation}

\subsection{Numerical model validation}
\begin{figure}[H]
\centering
\includegraphics[width=.9\textwidth]{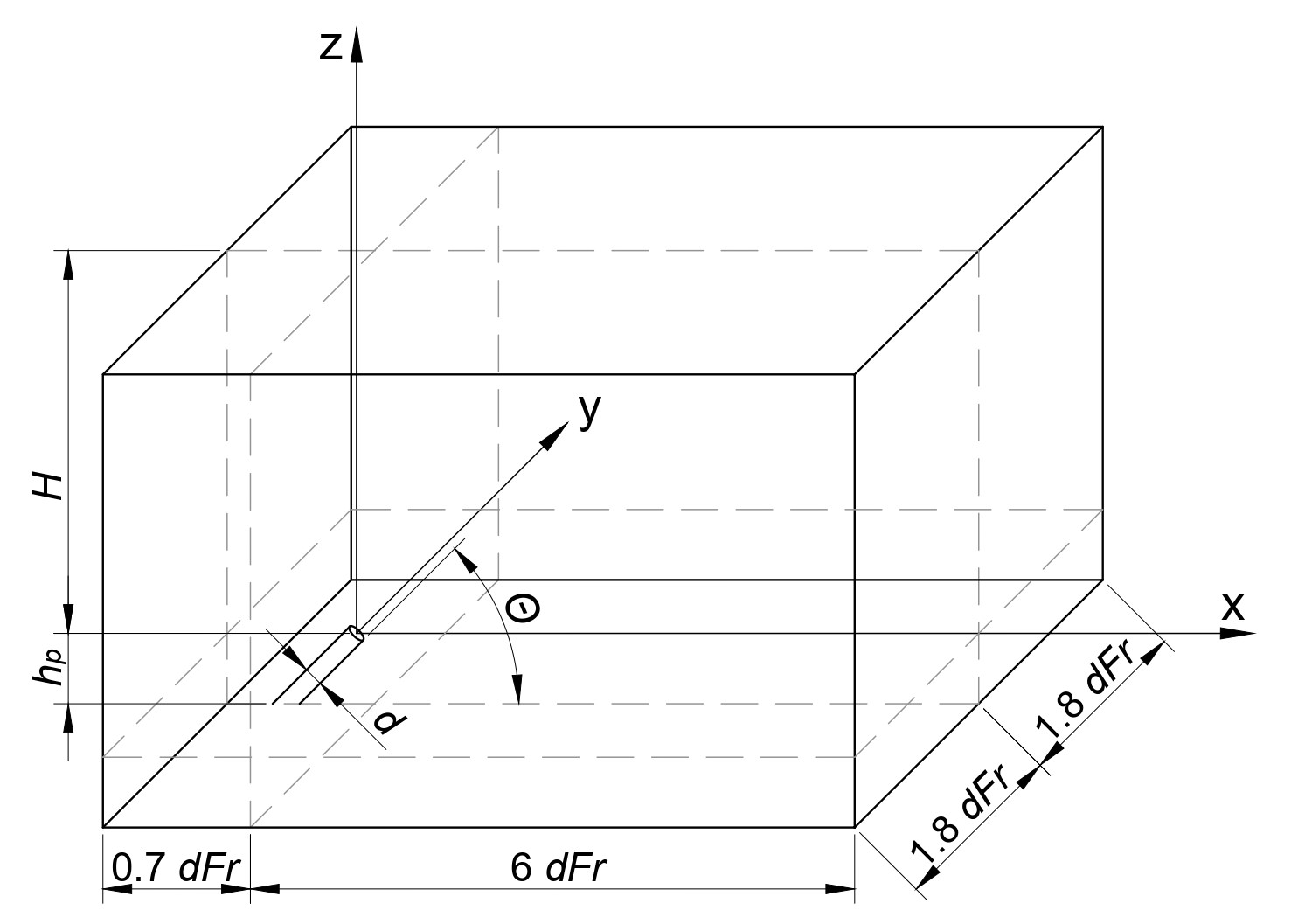}
\caption{Numerical domain.}
\label{fig:domain}
\end{figure}

To validate the numerical simulation of negatively inclined buoyant jets, experimental data from previous cases were used (\cite{zhang2016}), so numerical settings were set to reproduce experimental conditions. Dimensions of the domain are shown in Fig. \ref{fig:domain}. A nozzle diameter of 5.8 millimetres was angled at 45$^{\circ}$. Pipe height $h_p$ was 70 mm and the distance of the nozzle tip from the water surface $H$ was 150 mm. 
The origin of a coordinate system was set in the middle of the nozzle. CfMesh was used to create three meshes with different element numbers and refinement in the zones near the nozzle.

To model mixing and effluent behaviour of the jet, the OpenFOAM (\cite{Jasak07openfoam:a}) solver \textit{twoLiquidMixingFoam} was applied. It is based on solving the transient 3D Navier-Stokes equations for fluid flow and an additional phase advection-diffusion equation to represent the mixing of two fluids with different densities. The volume fraction of fluid is represented with an additional variable $\alpha$, which was set to zero in the initial moment throughout the whole domain, while at the pipe inlet, where the brine enters the domain, it was $\alpha$ = 1.

The back, front, right and left boundaries of the domain were represented with the Neumann boundary condition ($zeroGradient$ in OpenFOAM). 
The top surface was treated as a free slip boundary condition, while at the nozzle and bottom, the standard turbulence wall functions and the no slip condition were applied. Initial velocity in the domain was set to zero while inlet pipe velocity was set to 0.515 m/s. As suggested by the results from previous studies (\cite{gildeh2015numerical}), a realizable $k-\epsilon$ turbulence model with a turbulence intensity of 10\% was selected. 

Courant-Friedrichs-Lewy number (CFL) was set to be always less than 1, to adapt the time step and ensure the stability and convergence of simulation results. Effluent and ambient densities were set to 1034 kg/m$^3$ and 997 kg/m$^3$, respectively. Turbulent Schmidt number, kinematic viscosity and diffusivity were set as 0.7, 10$^{-6}$ m$^{2}$/s and 10$^{-9}$ m$^{2}$/s, respectively. The total simulation time is 120 seconds, while the results were time-averaged from 50 to 120 seconds, as in \cite{zhang2016}. 

For Laplacian terms and temporal discretization, the corrected Gauss linear scheme and second-order backward scheme were used. Gauss linear and Gauss VanLeer schemes were used for velocity and $\alpha$, while for the $k$ and $\epsilon$ Gauss upwind schemes were set. The convergence criterion for velocity and pressure was set as 10$^{-6}$. The preconditional conjugate gradient (PCG) method and diagonal incomplete Cholesky (DIC) preconditioner were used to solve the pressure field. For other fields including $U$, $k$ and $\epsilon$, preconditioned bi-conjugate gradient (PBiCGStab) method with diagonal incomplete LU (DILU) preconditioner was used. To reduce calculation time, simulation was performed in an HPC environment on the Bura supercomputer at the Center for Advanced Computing and Modeling (CNRM), University of Rijeka. The total execution time of simulation with medium mesh size performed on four nodes (96 CPU cores) was 22 hours.

To compare results and establish mesh convergence, a simulation of three different mesh sizes was calculated. Coarse, medium and fine mesh consist of $5.76\cdot 10^{5}$, $1.43\cdot 10^{6}$, $3.05\cdot 10^{6}$, respectively. Results of Grid Convergence Index (GCI) (\cite{roache1998verification}), presented in Table \ref{tab:gci_table}, are calculated for horizontal and vertical locations of the centerline peak $x_m$  and $z_m$. From the obtained data, it can be observed that with the increase in the number of elements, the values are converging with the asymptotic range i.e. $GCI_{m,f} / (r^p \cdot GCI_{c,m}) \sim 1$, thus using a medium mesh for comparison with experimental data is adequate.

\begin{table}[!ht]
\centering
\caption{GCI for time-averaged horizontal and vertical locations of the centerline peak $x_m$  and $z_m$.}
\label{tab:gci_table}
\centering
\begin{tabular}{lllllll}
Mesh type& Number of cells& Ratio& $x_{m}${[}m{]}& GCI$_{xm}${[}\%{]} & $z_{m}${[}m{]} & GCI$_{zm}${[}\%{]} \\ \hline
Coarse    & 675852          & -     & 0.11197    & -            & 0.0639     & -            \\ \hline
Medium    & 1431118         & 1.3   & 0.11893    & 0.07211      & 0.07153    & 0.03676      \\ \hline
Fine      & 3047282         & 1.3   & 0.12236    & 0.03396      & 0.07315    & 0.00746     
\end{tabular}
\end{table}

\begin{figure}[H]
\centering
\includegraphics[width=\textwidth]{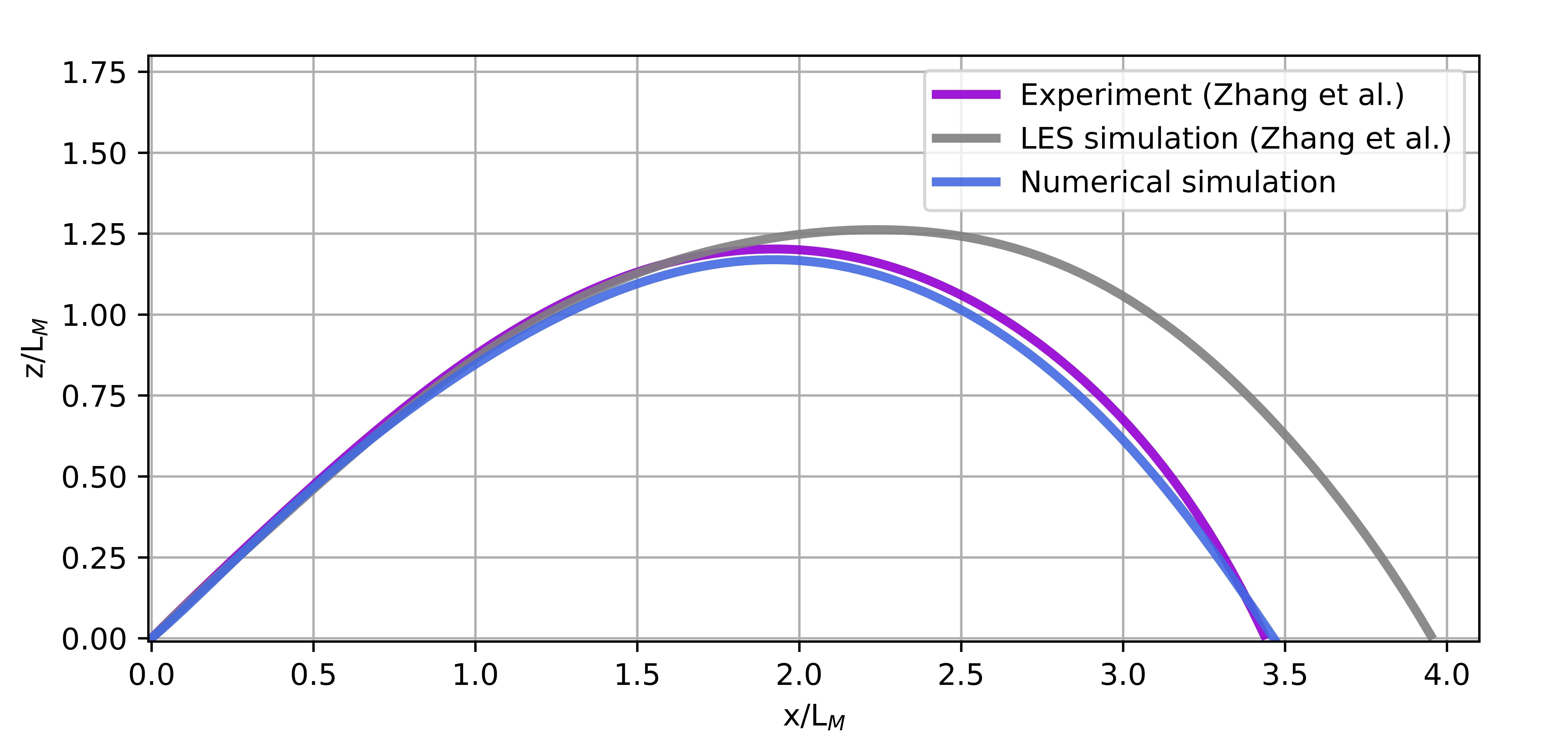}
\caption{Comparison of experimental, LES (\cite{zhang2016}) and simulation centerline results.}
\label{fig:compare}
\end{figure}

Fig. \ref{fig:compare} shows the results of time-averaged centerlines for experimental and LES simulation from the study by \cite{zhang2016} and numerical simulation results from this research. Results are calculated as maximum concentration points along the x direction. The value of root mean squared error (RMSE) between numerical results and experimental data is 0.00277, while the coefficient of determination (R$^2$) is 0.98475. Therefore, it can be concluded that a numerical model can very well predict experimental data.

\begin{figure}[H]
\centering
    \subfloat[]{\label{a}\includegraphics[width=.5\linewidth, clip]{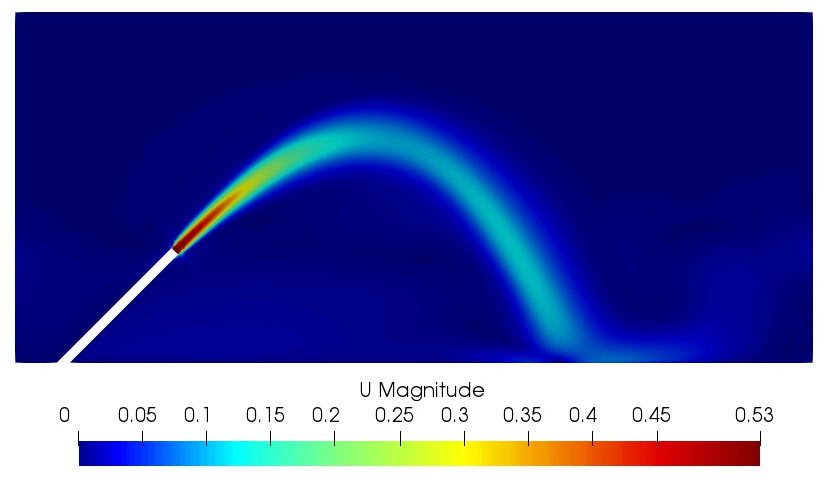}}\hfill
    \subfloat[]{\label{b}\includegraphics[width=.5\linewidth, clip]{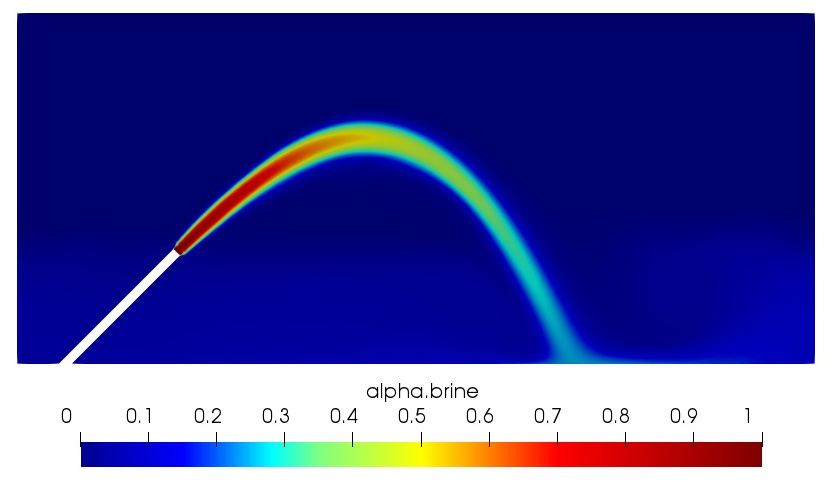}}
    \caption{Time averaged a) velocity (U magnitude) and b) concentration contours at the centre plane for validation case.}
   \label{fig:simulation_results}
\end{figure}

In Fig. \ref{fig:simulation_results}, the time-averaged velocity and concentration results are shown. It can be observed that the more distant the area is from the jet and centerline, the concentration and velocity rapidly decrease.

\subsection{ML model and feature selection}
Numerous CFD simulations were used to generate a dataset which was used to create an ML model. The input values were obtained by randomly assigning values (using the Python 3 programming language), however, several conditions had to be met. The density of the effluent had to be greater than the density of the water to create negative buoyant jets and the height of the water had to be at least twice the height of the pipe so that the centerline of the jet could not touch the surface. Furthermore, the set of input values was discarded if they had already been assigned.

Basic input data of ML models consist of flow parameters: angle $\theta$, water $H$ and pipe height $h_p$, nozzle diameter $d$, effluent velocity $U_0$, brine $\rho_b$ and ambient $\rho_a$ density. The interval of considered values for the basic input data is presented in Table \ref{tab:limit}, with a large range of Froude numbers from 1 to 94 and Reynolds numbers from 413 to 1390980. A total of 500 simulations with different flow conditions were calculated. Geometry was created based on Fig. \ref{fig:domain}. The mesh sizes for all cases ranged from $8\cdot 10^{5}$ to $2\cdot 10^{6}$. All other numerical settings were set the same as in the validation case.

\begin{table}[!ht] 
\caption{The minimum and maximum range for all ML model input variables.} 
\label{tab:limit}
\centering
\begin{tabular}{cccccccc}
        & $\theta$ ($^{\circ}$)& $h_p$ (m) & $H$ (m) & $d$ (m) & $U_0$ (m/s) &  $\rho_b$ (kg/m$^3$) &  $\rho_a$ (kg/m$^3$)   \\ \hline
Min         & 5        & 0.02    & 0.11 & 0.002 & 0.059   & 1000       & 980        \\ \hline
Max         & 80       & 5.89    & 9.8  & 0.39  & 4.96    & 1059       & 1030       \\ \hline
Average     & 51.2     & 1.39    & 5.04 & 0.15  & 2.37    & 1036       & 1006       \\ \hline
Std dev     & 14.5     & 1.02    & 2.03 & 0.09  & 1.31    & 14.36      & 11.52     
\end{tabular}
\end{table}

The ML model had a total of 5 output features, including return point location $x_r$, impact point location $x_i$, terminal rise height $z_t$, horizontal and vertical locations of the centerline peak $x_m$ and $z_m$. The range of output variables are presented in Table \ref{tab:limit2}. For ML prediction, the multi-target regression model was applied, with the relationship as follows: 

\begin{equation}\label{eq:3}
x_m, z_m, z_t, x_r, x_i = f(\theta, h_p, H, d, U_0, \rho_b, \rho_a )
\end{equation}

\begin{table}[!ht] 
\caption{The minimum and maximum range for ML model output variables.} 
\label{tab:limit2}
\centering
\begin{tabular}{cccccc}
& $x_{m}$ & $z_{m}$ & $z_{t}$ & $x_{r}$ & $x_{i}$        \\ \hline
Min     & 0.018   & 0.001   & 0.008   & 0.016   & 0.048  \\ \hline
Max     & 9.046   & 5.619   & 7.350   & 11.662  & 12.036 \\ \hline
Average & 2.042   & 1.514   & 2.267   & 3.676   & 4.194  \\ \hline
Std dev & 1.436   & 1.046   & 1.462   & 2.500   & 2.633 
\end{tabular}
\end{table}

\begin{figure}[H]
\centering
\includegraphics[width=\textwidth]{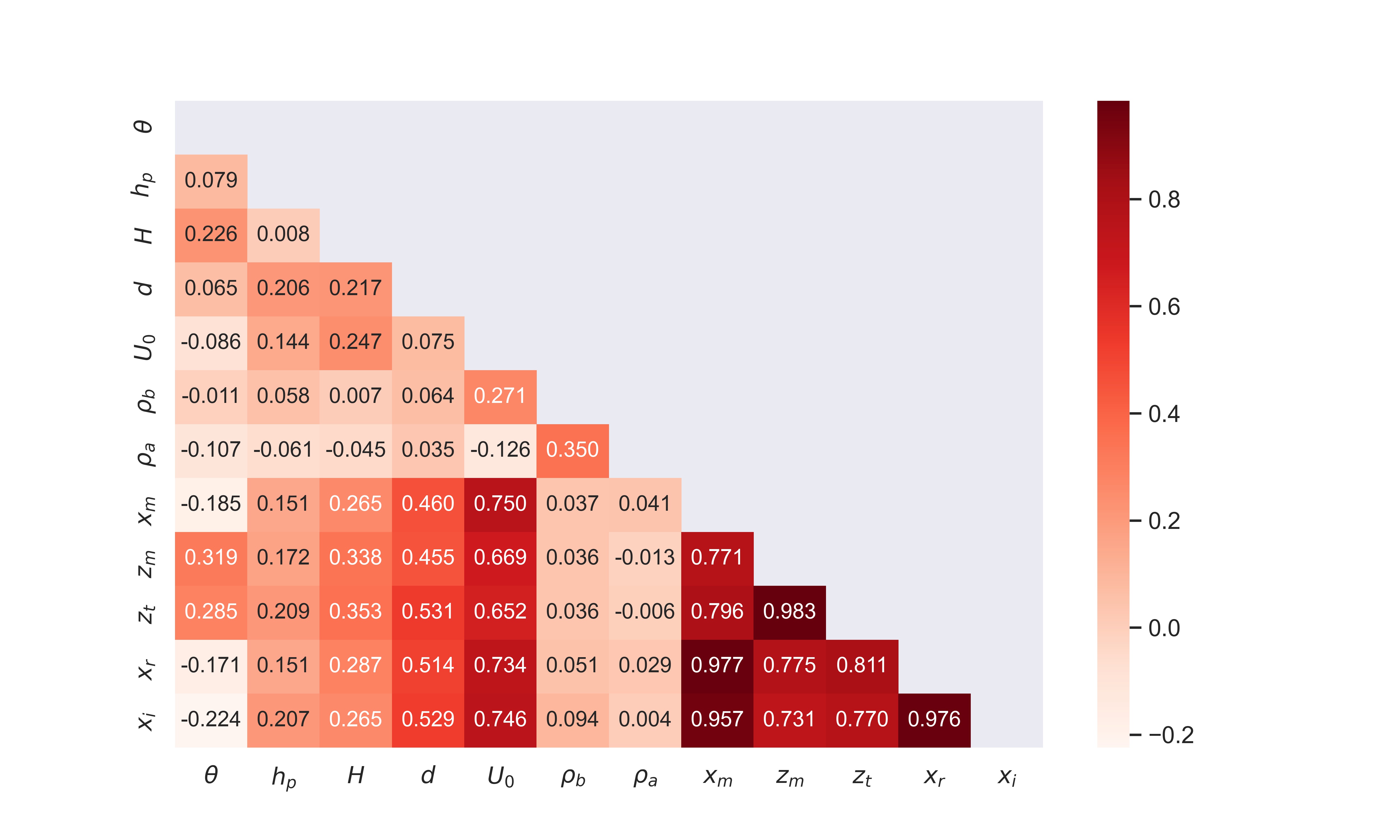}
\caption{Correlation heatmap displaying the relationship between the features based on Pearson's correlation coefficient.}
\label{fig:heatmap}
\end{figure}

Furthermore, Pearson’s correlation matrix was calculated for input and output data and is shown in Fig. \ref{fig:heatmap} to exhibit the linear relationship strength between variables. It can be concluded that all input variables have a positive correlation except the $\theta$ and $\rho_a$. According to this graph, $U_0$ has the greatest influence, but in order to obtain a more detailed analysis, the SHAP feature interpretation method will be applied.

\begin{figure}[H]
\centering
\includegraphics[width=\textwidth]{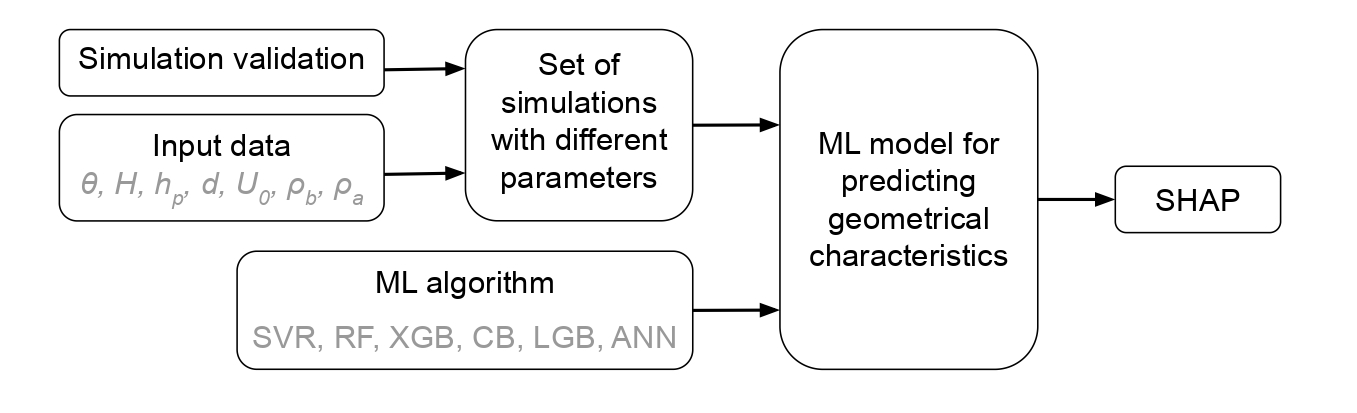}
\caption{ML model flowchart which includes data inputs, algorithms and the ML model output.}
\label{fig:diagram}
\end{figure}

In Fig. \ref{fig:diagram} a flowchart for creating the ML model is shown. Firstly, after a numerical model validation, simulations with different input variables were obtained. Secondly, an array of machine learning algorithms were used for ML model training based on the data. The final ML model is trained and validated to predict the geometrical characteristics of negatively buoyant jets (based on Equation \ref{eq:3}).

\subsection{Machine learning algorithms}

Different machine learning models have been applied for predicting geometrical characteristics for negatively inclined buoyant jets to give a full insight into the capabilities and best prediction performance of the different models.

Support Vector Regression (SVR) is a variant of support vector machines (SVM) for regression problems (\cite{drucker1996support}). The main principle is mapping input features to high-dimensional space separated by a hyperplane to minimize error and maximize the margin using a kernel function. 

Tree-based ensemble methods are a group of machine learning algorithms based on combining several decision trees to produce better performances than a single decision tree. 
The most popular tree-based ensemble method is Random Forests (RF) which applies averaging on several estimators built independently to improve prediction and control overfitting (\cite{breiman2001random}). 
Boosting methods are a group of tree-based ensemble methods which build estimators sequentially, along with the reduction of bias from combined estimators. 
XGBoost (XGB) is a scalable machine learning system for tree boosting proposed by \cite{chen2016xgboost}. It utilises a sparsity-aware algorithm and weighted quantile sketch procedure for approximate learning.
Catboost (CB) is another type of tree boosting method established on ordered boosting, permutation-driven alternative and a new algorithm for processing categorical features (\cite{prokhorenkova2018catboost}). 
Another boosting method used is LightGBM (LGB) (\cite{ke2017lightgbm}) which includes two novel techniques, Gradient-based One-Side Sampling (GOSS) to find the best split value and Exclusive Feature Bundling (EFB) to reduce the feature space complexity.  

Finally, an Artificial Neural Network (ANN) is a machine learning model made according to the structure of a biological neural network. Most of the previous studies used ANN for the prediction of various parameters in desalination systems (\cite{behnam2022review}). In this paper, Multi-Layer Perceptron (MLP) Artificial Neural Network (ANN) is applied. The basic structure consists of three types of layers: input, hidden, and output with neurons interconnected with links. Each neuron has an activation function calculated from input data resulting in weights, additionally adjusted based on the accuracy of prediction (\cite{nielsen2015neural}).

All algorithms were implemented in Python 3.8. programming language. SVR, RF and ANN implementations in scikit-learn 1.1.1 Python module (\cite{pedregosa2011scikit}) were used, CB from the catboost 1.0.0 module, XGB from the xgboost 1.6.0 module and LGB from lightgbm 3.3.2. 
The input data for SVR and ANN were standardized between 0 and 1 using MinMaxScaler function from the scikit-learn module. This standardization procedure did not affect the tree-based ensemble methods including RF, XGB, CB and LGB. 
To build a model for all geometrical characteristics simultaneously, the MultiOutputRegressor function implemented in the scikit-learn model was used for SVR, RF, CB and LGB. 
XGB and ANN support multi-output regression, so an additional function is not needed. Hyperparameters of all ML algorithms were tuned using a cross-validated grid search algorithm over a parameter grid, and are presented in Table \ref{tab:hyper}. Other parameters were set as default.

\begin{table}[!ht]
\caption{Hyperparameters for ML models.}
\label{tab:hyper}
\centering
\begin{tabular}{ll}
ML model & Hyperparameters                                                            \\ \hline
SVR      & Kernel: rbf, $C$: 1000, $\epsilon$: 0.005, $\gamma$: 0.5                             \\
RF       & Estimators: 250, maximum depth: 90                                             \\
XGB      & Estimators: 300, maximum depth: 2                                              \\
CB       & Estimators: 300, maximum depth: 5                                              \\
LGB      & Estimators: 250, maximum depth: 25, number of leaves: 7                               \\
ANN      & Hidden layers: 3, neurons 25, solver: lbfgs, activation: tanh, $\alpha$=0.05
\end{tabular}
\end{table}

For comparison of predicted ($y_{i, predicted}$) and observed ($y_{i, observed}$) results, the root mean squared error (RMSE) and coefficient of determination (R$^2$) were used, which can be calculated according to: 
\begin{equation}\label{eq:4}
RMSE = \sqrt{\frac{1}{n}\sum_{i=1}^{n}[(y_{i, observed}-y_{i, predicted})]^2}
\end{equation}

\begin{equation}\label{eq:5}
R^2 = 1-\frac{\sum_{i=1}^{n}[(y_{i, observed}-y_{i, predicted})]^2}{\sum_{i=1}^{n}[(y_{i, observed}-\overline{y})]^2}
\end{equation}

where $\overline{y}$ and $n$ represent the average of observed data and the number of data points.

\subsection{Shapley additive explanations}
SHapley Additive exPlanations (SHAP) is a game theoretic-based method for interpreting feature importance from machine learning models (\cite{lundberg2017unified}). In contrast to Pearson's correlation matrix, which assessed only a single linear relationship between two features, SHAP assessed both linear and nonlinear multivariable relationships, giving full insight into variable importance. SHAP values were calculated using shap 0.39.0 Python module based on a multi-output ANN model using the KernelExplainer function.

\section{Result and discussion}
\subsection{Results analysis}

In order to rigorously validate the ML models, the dataset was randomly split 90-10 for train$\backslash$test and validation.
Firstly, on the train$\backslash$test dataset, average and standard deviation were calculated for a K-fold cross validation procedure with K=5 with results presented in Table \ref{tab:result1}.
Secondly, the 90\% train dataset was used to train the model, and the 10\% dataset was used for model validation, with results presented in Table \ref{tab:result2}. The bolded value indicates the best result for each geometric characteristic. It can be observed that the best results are obtained using the ANN algorithm. Tree-based ensemble methods predict results with lower accuracy, especially RF. The best results from tree-based ensemble models are obtained by CB.

\begin{table}[H]
\caption{Results of R$^2$ and RMSE with standard deviation (std) of ML models on train$\backslash$test dataset for K-fold cross validation.}
\label{tab:result1}
\centering
\begin{adjustbox}{width=1\textwidth}
\begin{tabular}{cccccccc}
                         &       & SVR         & RF          & XGB         & CB          & LGB         & ANN         \\ \hline
\multirow{2}{*}{$x_{m}$} & R$^2$$\pm$std & 0.944$\pm$0.016 & 0.883$\pm$0.418 & 0.911$\pm$0.027 & 0.932$\pm$0.011 & 0.923$\pm$0.022 & \textbf{0.955$\pm$0.011} \\ \cline{2-8} 
                         & RMSE$\pm$std  & 0.337$\pm$0.035 & 0.481$\pm$0.055 & 0.423$\pm$0.069 & 0.365$\pm$0.019 & 0.397$\pm$0.059 & \textbf{0.298$\pm$0.053} \\ \hline
\multirow{2}{*}{$z_{m}$} & R$^2$$\pm$std & 0.962$\pm$0.004 & 0.855$\pm$0.007 & 0.917$\pm$0.024 & 0.928$\pm$0.026 & 0.925$\pm$0.013 & \textbf{0.971$\pm$0.004} \\ \cline{2-8} 
                         & RMSE$\pm$std  & 0.208$\pm$0.022 & 0.393$\pm$0.025 & 0.295$\pm$0.027 & 0.278$\pm$0.052 & 0.281$\pm$0.022 & \textbf{0.176$\pm$0.012} \\ \hline
\multirow{2}{*}{$z_{t}$} & R$^2$$\pm$std & 0.956$\pm$0.009 & 0.861$\pm$0.010 & 0.926$\pm$0.012 & 0.941$\pm$0.014 & 0.926$\pm$0.016 & \textbf{0.971$\pm$0.007} \\ \cline{2-8} 
                         & RMSE$\pm$std  & 0.311$\pm$0.021 & 0.537$\pm$0.036 & 0.394$\pm$0.012 & 0.352$\pm$0.052 & 0.380$\pm$0.039 & \textbf{0.246$\pm$0.037} \\ \hline
\multirow{2}{*}{$x_{r}$} & R$^2$$\pm$std & 0.959$\pm$0.013 & 0.894$\pm$0.026 & 0.921$\pm$0.021 & 0.941$\pm$0.012 & 0.930$\pm$0.006 & \textbf{0.973$\pm$0.006} \\ \cline{2-8} 
                         & RMSE$\pm$std  & 0.501$\pm$0.069 & 0.825$\pm$0.112 & 0.702$\pm$0.095 & 0.590$\pm$0.086 & 0.656$\pm$0.048 & \textbf{0.402$\pm$0.049} \\ \hline
\multirow{2}{*}{$x_{i}$} & R$^2$$\pm$std & 0.946$\pm$0.014 & 0.902$\pm$0.011 & 0.915$\pm$0.020 & 0.941$\pm$0.020 & 0.935$\pm$0.010 & \textbf{0.956$\pm$0.009} \\ \cline{2-8} 
                         & RMSE$\pm$std  & 0.606$\pm$0.067 & 0.826$\pm$0.071 & 0.761$\pm$0.065 & 0.619$\pm$0.076 & 0.673$\pm$0.083 & \textbf{0.540$\pm$0.019}
\end{tabular}
\end{adjustbox}
\end{table}

\begin{table}[H]
\caption{Results of R$^2$ and RMSE of ML models for validation.}
\centering
\begin{adjustbox}{width=1\textwidth}
\label{tab:result2}
\begin{tabular}{ccccccccccccc}
        & \multicolumn{2}{c}{SVR} & \multicolumn{2}{c}{RF} & \multicolumn{2}{c}{XGB} & \multicolumn{2}{c}{CB} & \multicolumn{2}{c}{LGB} & \multicolumn{2}{c}{ANN}  \\ \hline
        & R$^2$      & RMSE       & R$^2$      & RMSE      & R$^2$      & RMSE       & R$^2$      & RMSE      & R$^2$      & RMSE       & R$^2$      & RMSE        \\ \hline
$x_{m}$ & 0.960      & 0.291      & 0.890      & 0.488     & 0.924      & 0.374      & 0.936      & 0.352     & 0.924      & 0.413      & \textbf{0.963}      & \textbf{0.259}      \\ \hline
$z_{m}$ & 0.971      & 0.160      & 0.868      & 0.331     & 0.930      & 0.277      & 0.932      & 0.241     & 0.948      & 0.241      & \textbf{0.980}      & \textbf{0.156}       \\ \hline
$z_{t}$ & 0.962      & 0.255      & 0.880      & 0.498     & 0.926      & 0.390      & 0.957      & 0.271     & 0.935      & 0.364      & \textbf{0.984}      & \textbf{0.200}       \\ \hline
$x_{r}$ & 0.969      & 0.482      & 0.910      & 0.715     & 0.934      & 0.644      & 0.951      & 0.557     & 0.948      & 0.679      & \textbf{0.987}      & \textbf{0.290}       \\ \hline
$x_{i}$ & 0.951      & 0.635      & 0.902      & 0.853     & 0.924      & 0.737      & 0.951      & 0.587     & 0.935      & 0.652      & \textbf{0.962}      & \textbf{0.491}       \\ \hline
Average & 0.962      & 0.364      & 0.890      & 0.577     & 0.928      & 0.484      & 0.945      & 0.402     & 0.938      & 0.470      & \textbf{0.975}      & \textbf{0.279}    
\end{tabular}
\end{adjustbox}
\end{table}

\begin{figure}[H]
\centering
\includegraphics[width=\textwidth]{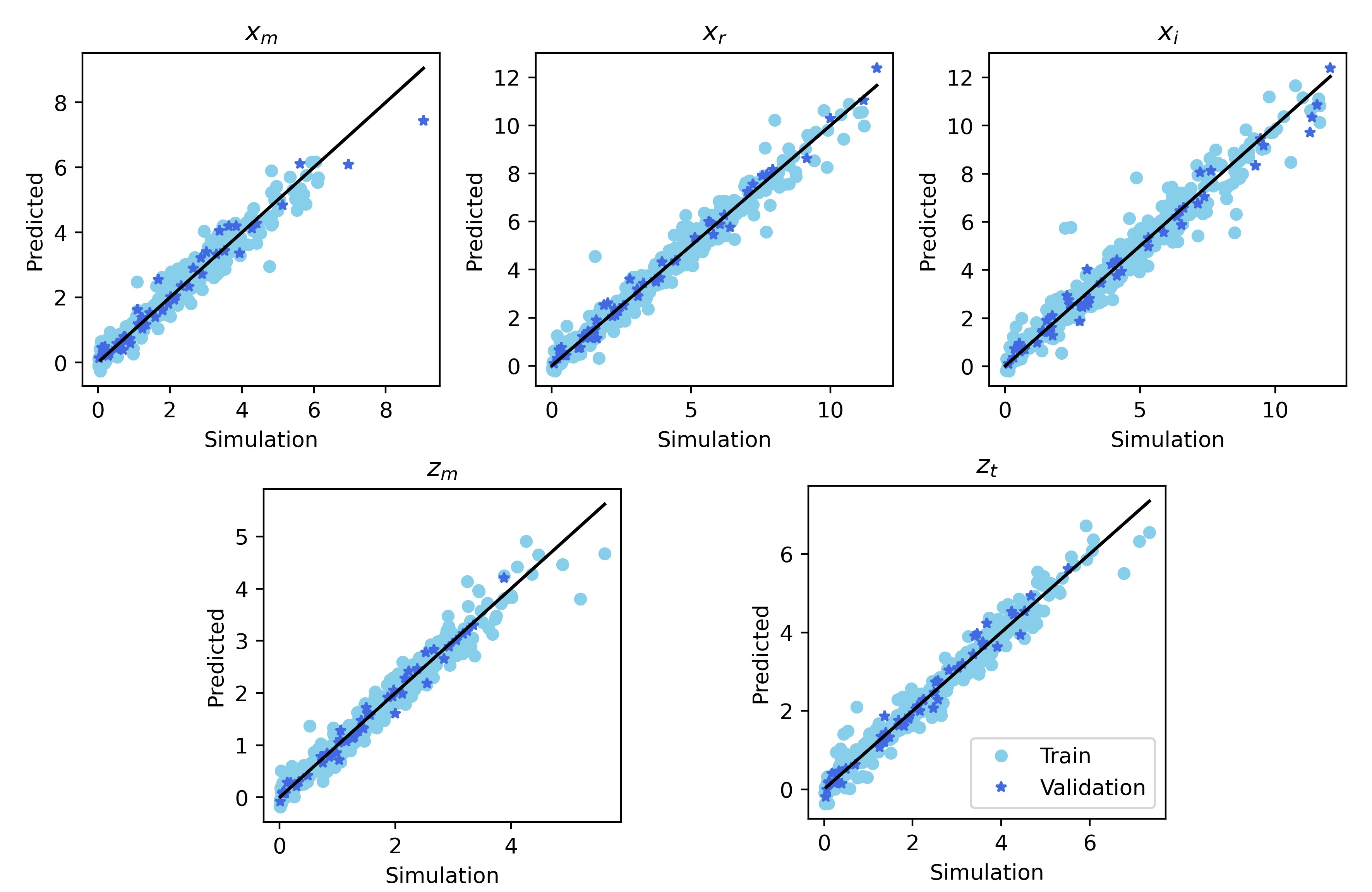}
\caption{Comparison of predicted and simulated data for ANN ML model.}
\label{fig:predsim}
\end{figure}

Considering that ANN gave the best results for the prediction of all geometrical characteristics, it is used for further detailed investigation. In Fig. \ref{fig:predsim}, a comparison between simulated and predicted results is shown. It can be observed that most of the train and validated data are located near the 1:1 line, indicating well model performance for all geometrical characteristics.

\begin{figure}[H]
\centering
\includegraphics[width=0.8\textwidth]{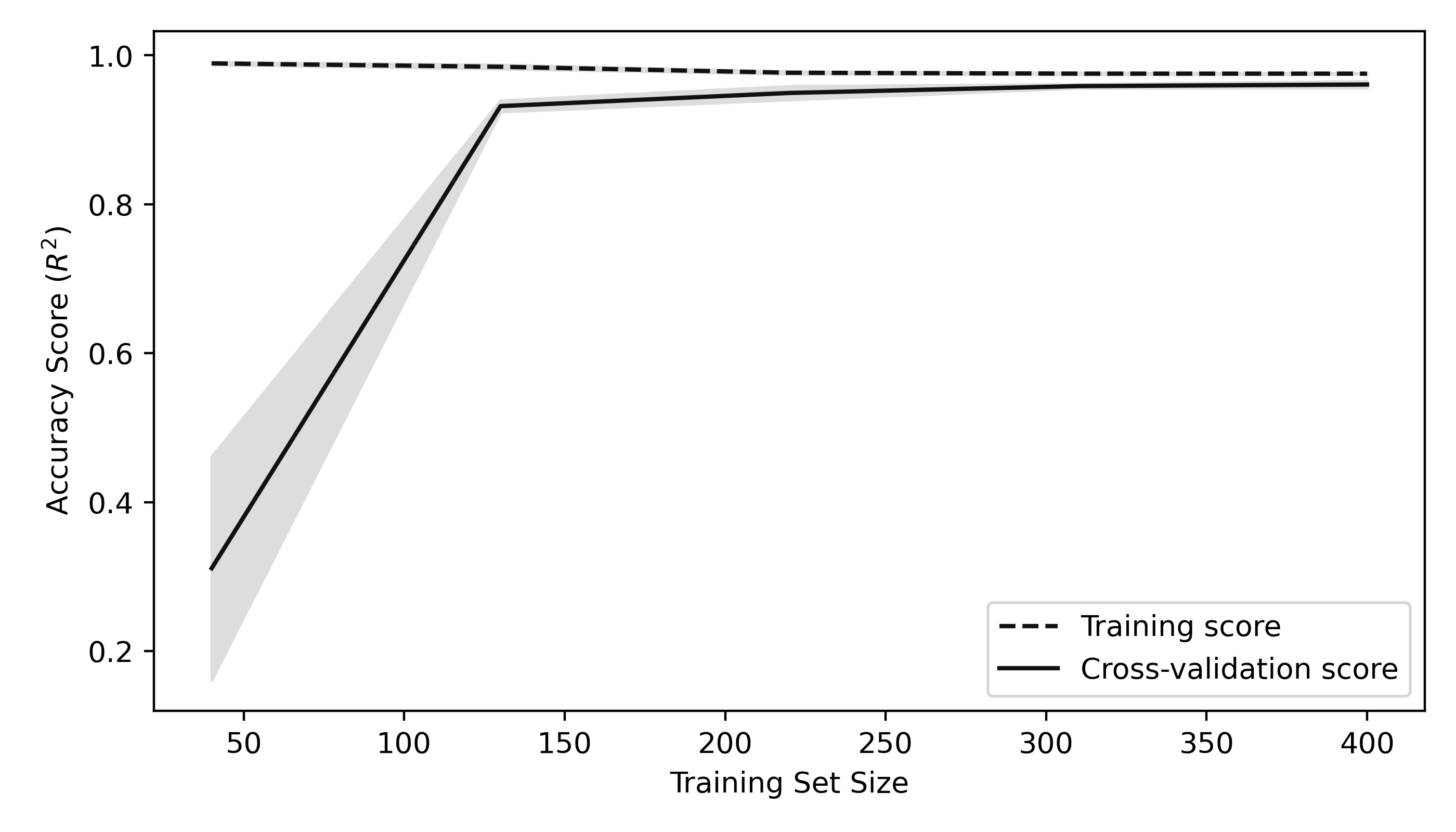}
\caption{The learning curve for all output data and ANN ML model.}
\label{fig:learncurve}
\end{figure}

Fig. \ref{fig:learncurve} shows the learning curve for the  ANN model. It represents the dependency between cross validated R$^2$ and training/test size. The shaded part shows the standard deviation of the results. It can be observed, that accuracy of K-fold cross validation is growing significantly until the training set size reaches 150, while further, a stagnant growth is achieved at around 300. For the whole dataset, training and cross-validation score are converged. Therefore, it can be concluded that this size of the dataset is sufficient for analysis and it is not necessary to generate more data because the accuracy of the model would not increase.

\subsection{SHAP feature interpretation}
\begin{figure}[H]
\centering
\includegraphics[width=\textwidth]{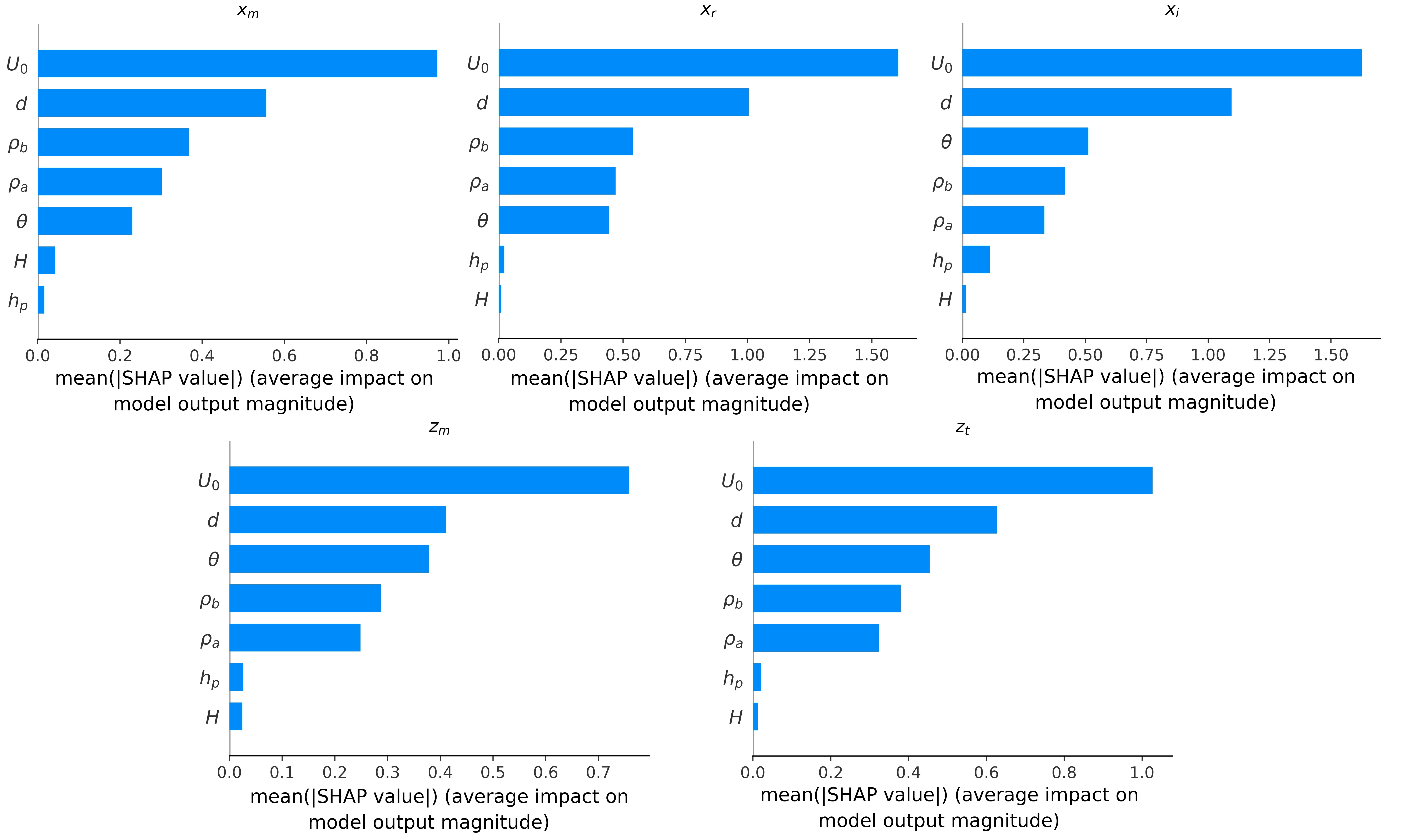}
\caption{SHAP features an important bar plot for geometrical characteristics.}
\label{fig:shapbarplot}
\end{figure}

In Fig. \ref{fig:shapbarplot} SHAP feature importance bar plots are presented. The input features are ranked by their importance in descending order based on their contribution.
The highest importance has $U_0$, followed by $d$ as in Fig. \ref{fig:heatmap} for all geometrical characteristics. Other SHAP variable importance order is different from Pearson correlation, implying the nonlinear relationship between variables.

\begin{figure}[H]
\centering
\includegraphics[width=\textwidth]{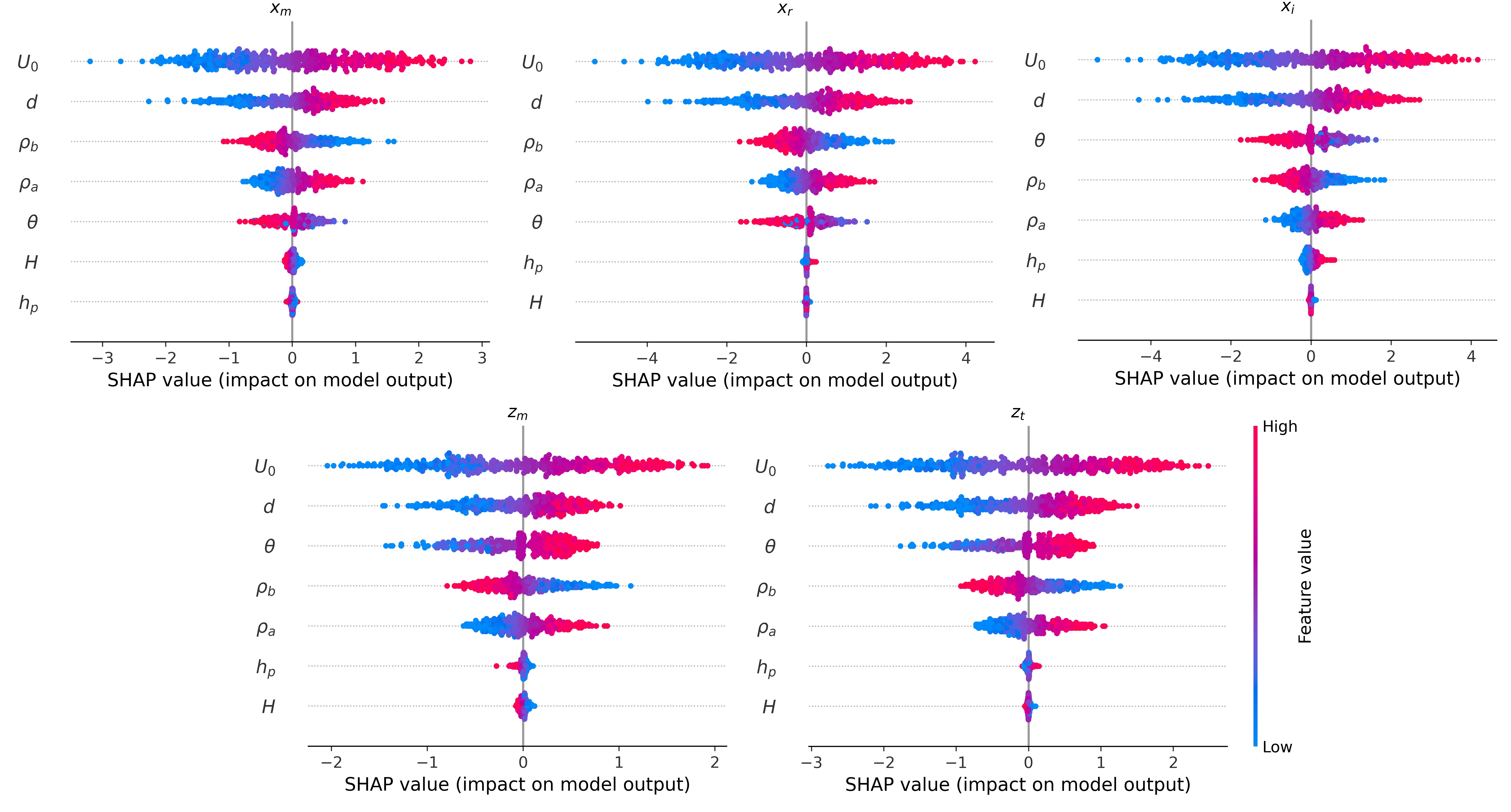}
\caption{SHAP summary plot on geometrical characteristics.}
\label{fig:shapsummary}
\end{figure}

To get the full insight of input feature influence, in Fig. \ref{fig:shapsummary}, a SHAP summary plot which shows the positive and negative relationships with geometrical characteristics is presented. On the x axis, feature contribution on the ML model output is presented. The colour of the dots presents the value of each observation, red implies higher values of input characteristics, while blue lower ones. Most of the input parameters show a positive correlation, which means that by increasing e.g. $U_0$, the value of geometrical characteristics will also be increased. A negative correlation can be observed for $\rho_b$ for all geometrical characteristics. Insignificantly low SHAP values are recorded for $H$ and $h_p$ indicating the small importance of geometrical characteristics.

\begin{figure}[H]
\centering
\includegraphics[width=\textwidth]{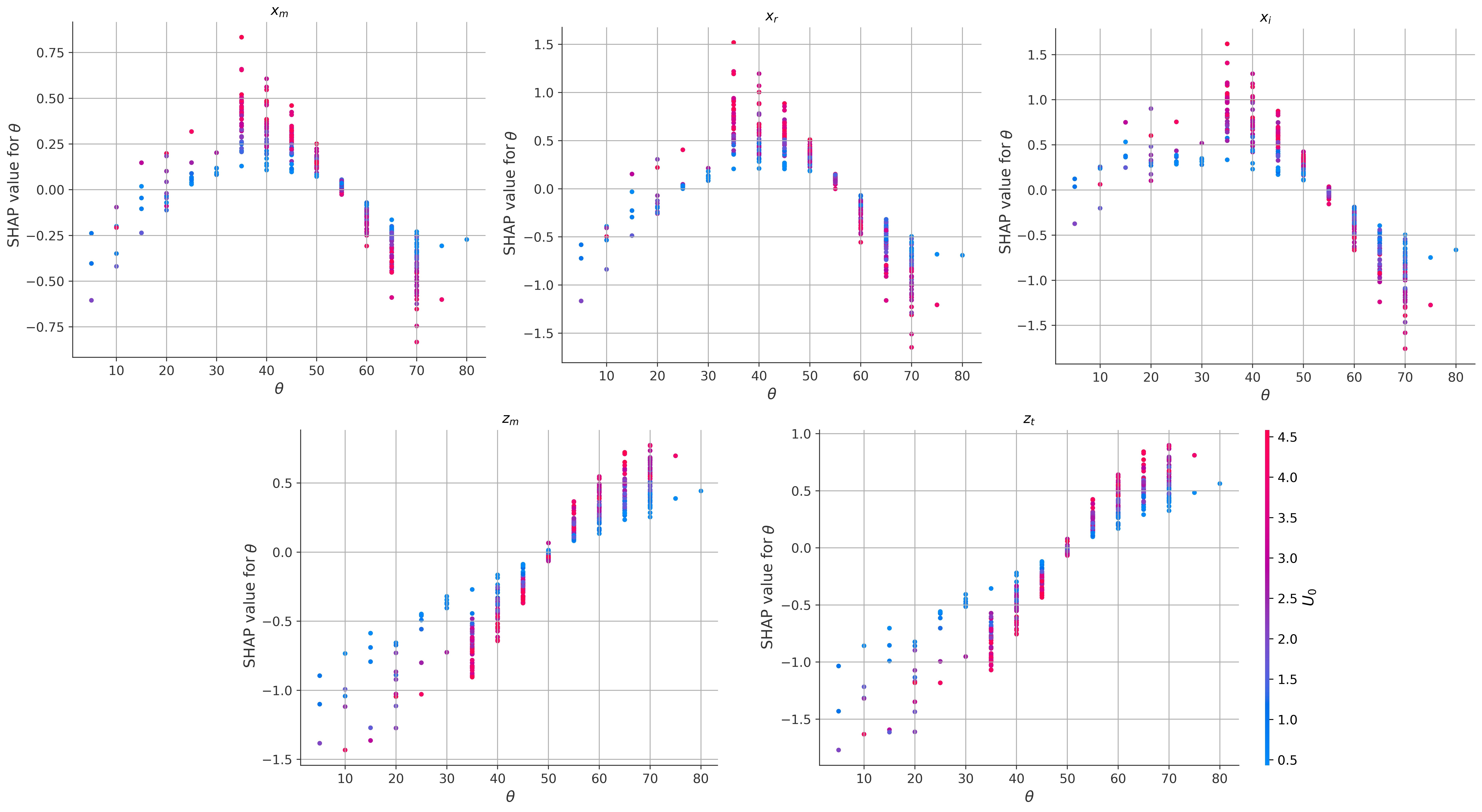}
\caption{SHAP partial dependence plots for $\theta$ and $U_0$.}
\label{fig:shapdepp}
\end{figure}

In Fig. \ref{fig:shapdepp}, a more detail influence of $\theta$ and $U_0$ on SHAP values is presented. Many previous works have discussed which angle is the best to place the desalination nozzle in order to get the best performances, mostly mentioning angles of 45$^{\circ}$ and 60$^{\circ}$. According to the upper images, a horizontal distance of geometrical characteristics will be most influenced by the 35$^{\circ}$, while vertical distance will increase with the angle. For 50$^{\circ}$ the impact of velocity will be the smallest. Therefore, it is not possible to provide only one best angle, but other flow parameters including $d$ and densities should be taken into account.

\subsection{SHAP analysis of Coanda and Shallow Water Effects}

As can be observed from Fig. \ref{fig:shapbarplot} and \ref{fig:shapsummary} water and pipe heights have the smallest importance values among all input variables summarized for the entire dataset. To observe specific cases, SHAP waterfall diagrams are plotted for one prediction result to identify if the impact of $h_p$ and $H$ increases in some flow conditions. Diagrams represent the positive and negative SHAP values, similar to points in Fig. \ref{fig:shapsummary}. However, due to the scale and general influence on the summary plot, it is not possible to investigate specific cases without waterfall diagrams.

\begin{figure}[H]
\centering
\includegraphics[width=\textwidth]{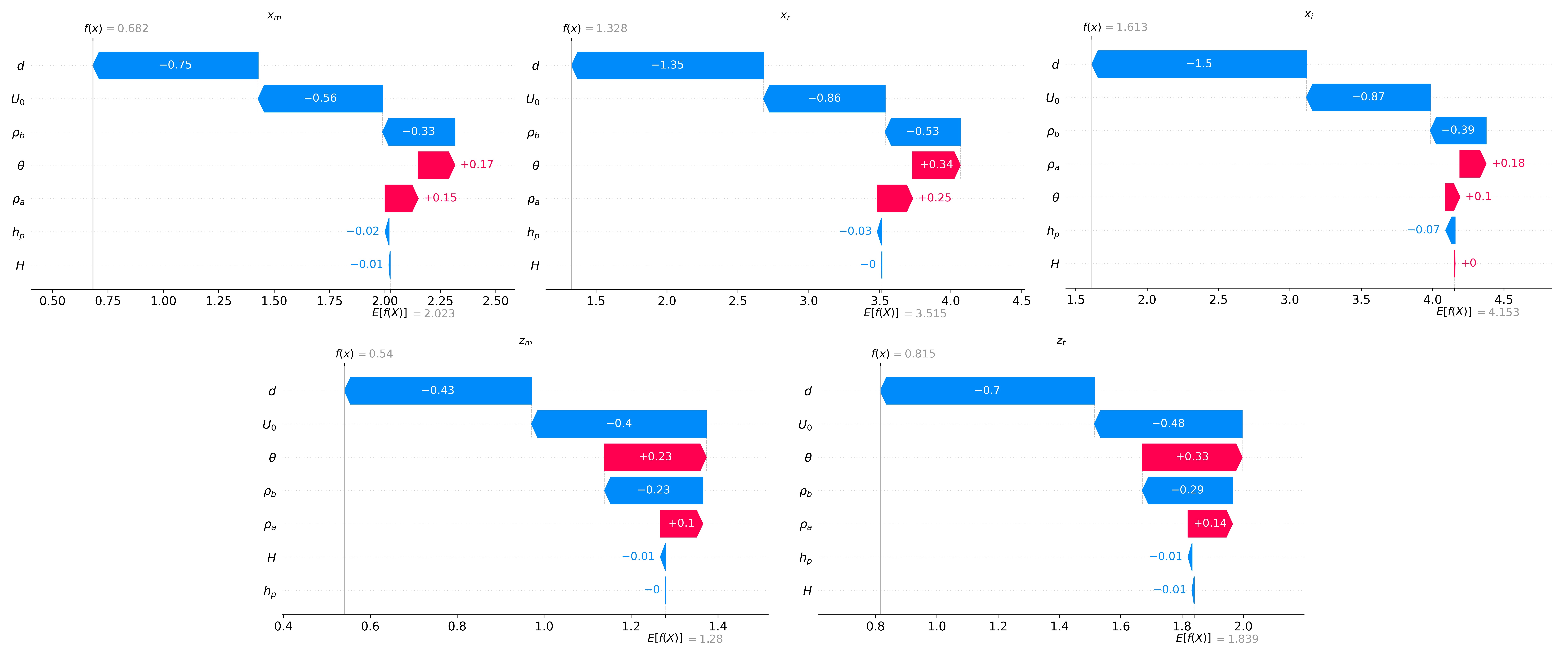}
\caption{SHAP waterfall plots case with $h_p$/$L_M$= 0.038 and 50$^{\circ}$. At the bottom of the plot, the value for  E(f(x)) represents the average predicted score for each output data of the dataset, while on the top, f(x) presents the predicted value of the output variables for this specific case.}
\label{fig:waterfall1}
\end{figure}

For Coanda effect, the controlling parameter by \cite{shao2010mixing} and \cite{ramezani2021effect} is $h_p$/$L_M$. Results from previous research presented a strong effect for $h_p$/$L_M$<0.2 and an angle of 30$^{\circ}$. For the angle of 45$^{\circ}$, the effect was weaker. Therefore, to understand the effect of the bed proximity and the Coanda effect more clearly, the SHAP waterfall diagram presented in Fig. \ref{fig:waterfall1} is used. Results are presented for smaller values in the dataset $h_p$/$L_M$= 0.038 and 50$^{\circ}$. It can be observed that $d$ and $U_0$ have the greatest influence on the results, while $h_p$ has a bigger influence than in previous results (Fig. \ref{fig:shapbarplot} and \ref{fig:shapsummary}) with overall results.

Previous literature (\cite{shao2010mixing}) indicates a larger influence of the Coanda effect on the horizontal variables, translating them downstream, while there was no influence on the vertical variables, which can be confirmed from these results. $\theta$ and $\rho_a$ show a positive impact on prediction, while other variables have a negative. However, this influence is still small and in order to increase the influence, $L_M$ should be further reduced to allow for $h_p$ to have a greater impact on the output results, which is the opposite of the desired environmental effect.

For the shallow water effect, the most important parameter is $d$ $Fr$/$H$ (\cite{abessi2016dense}). The effect occurs when the value of $d$ $Fr$/$H$ equals 0.95, 0.65 for inclinations, 30$^{\circ}$, 45$^{\circ}$, respectively (\cite{jiang2014mixing}). Larger angles yield a stronger shallow water effect.

\begin{figure}[H]
\centering
\includegraphics[width=\textwidth]{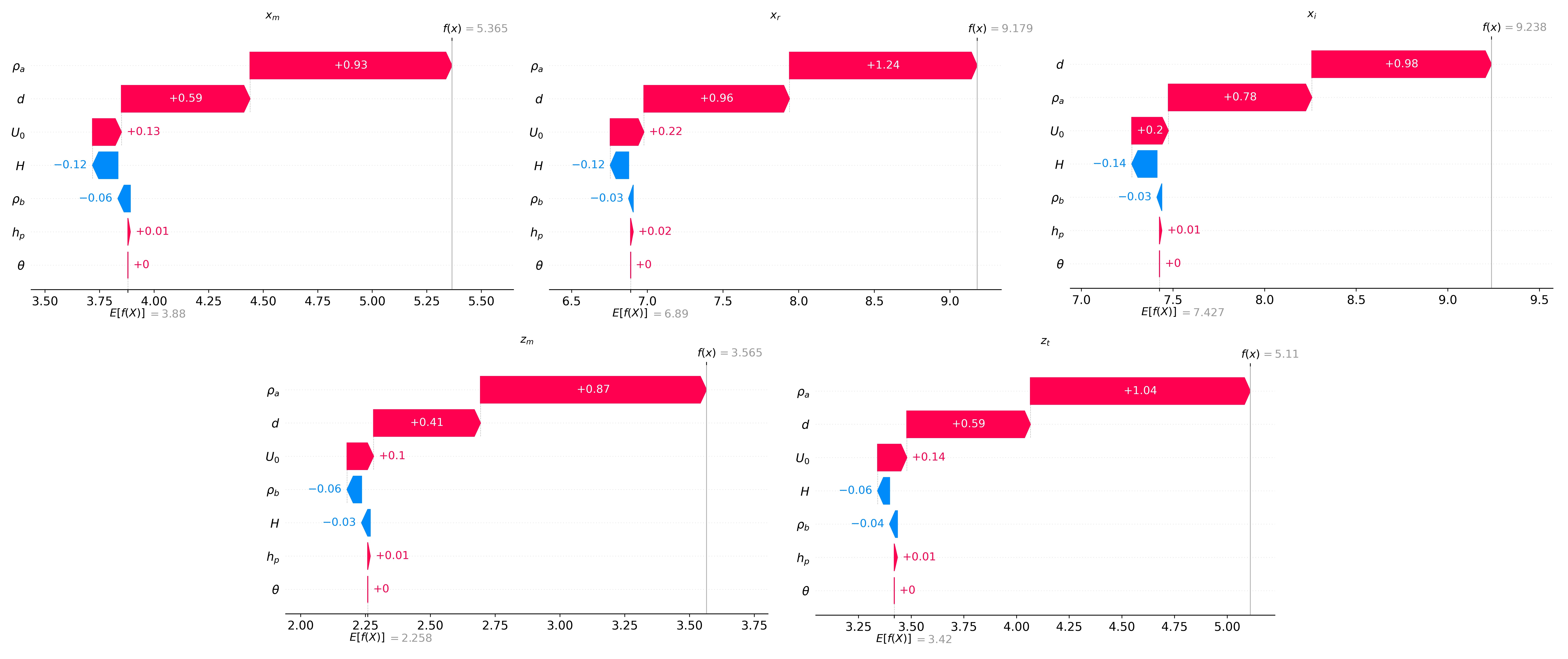}
\caption{SHAP waterfall plots case with $d$ $Fr$/($H$)= 0.65 and 45$^{\circ}$.}
\label{fig:waterfall2}
\end{figure}

To investigate this behaviour, the case with $d$ $Fr$/$H$=0.66 and 45$^{\circ}$ is investigated and presented in Fig. \ref{fig:waterfall2}. For this value, there is a change from the full submergence regime to the plume contact regime. In contrast to previous results (Fig. \ref{fig:shapsummary}), $H$ has a much greater influence. The biggest influence is on $x_m$, $x_r$ and $x_i$, moving them downstream due to surface attachment, as expected from previous literature (\cite{jiang2014mixing}). In contrast to the Coanda effect, the shallow water effect creates a negative influence of $H$, i.e. its reduction increases the value of the output variable. For the angle of 45$^{\circ}$, SHAP analysis gave the least importance to $\theta$, indicating that the angle has no positive or negative impact on results.

\subsection{Analysis based on previous research}

Results from previous research have been analyzed using the SHAP method and compared with CFD simulations. Flow parameters of the experiments and the result of geometrical characteristics from several studies were used (\cite{cipollina2005bench}, \cite{abessi2016dense}, \cite{azizi2020experimental}, \cite{jiang2014mixing}). To the authors' knowledge, this dataset contains all available data from previous experiments which hold all input information used in this ML model (most of the literature omits values such as water and pipe height or densities). 

Each researcher studied different geometrical characteristics, so the number of data for each output value is different and presented in Table \ref{tab:combined_database}. Since most of those points include values with inclination 30$^{\circ}$, 45$^{\circ}$ and 60$^{\circ}$ and the same values of some parameters such as $\rho_{a}$, for ML models and SHAP analysis, the data from those experiments were mixed with CFD simulation data.

Results were calculated on the same principle as in Tables \ref{tab:result1} and \ref{tab:result2}, splitting data on 90-10 for train$\backslash$test and validation, along with a 5-fold cross validation. The number of variables is different, therefore, in this case, the multioutput regression function was not utilized. For calculation, the same tuned ANN model was used as in the previous analysis. Similar results in Table \ref{tab:combined_database} can be observed as in Tables \ref{tab:result1} and \ref{tab:result2}. 

\begin{table}[!ht]
\centering
\caption{Results of ML models for a large dataset including previous experimental data.}
\label{tab:combined_database}
\begin{tabular}{cccccc}
        & Number of       & \multicolumn{2}{c}{R$^2$} & \multicolumn{2}{c}{RMSE} \\ \cline{3-6} 
        & additional data & K-fold         & Val.     & K-fold         & Val.    \\ \hline
$x_{m}$ & 130             & 0.965$\pm$0.007    & 0.977    & 0.278$\pm$0.022    & 0.203   \\ \hline
$z_{m}$ & 44              & 0.972$\pm$0.004    & 0.977    & 0.177$\pm$0.009    & 0.185   \\ \hline
$z_{t}$ & 80              & 0.977$\pm$0.002    & 0.981    & 0.232$\pm$0.016    & 0.215   \\ \hline
$x_{r}$ & 124             & 0.981$\pm$0.006    & 0.988    & 0.355$\pm$0.055    & 0.301   \\ \hline
$x_{i}$ & 88              & 0.971$\pm$0.006    & 0.977    & 0.470$\pm$0.053    & 0.471  
\end{tabular}
\end{table}

\begin{figure}[H]
\centering
\includegraphics[width=\textwidth]{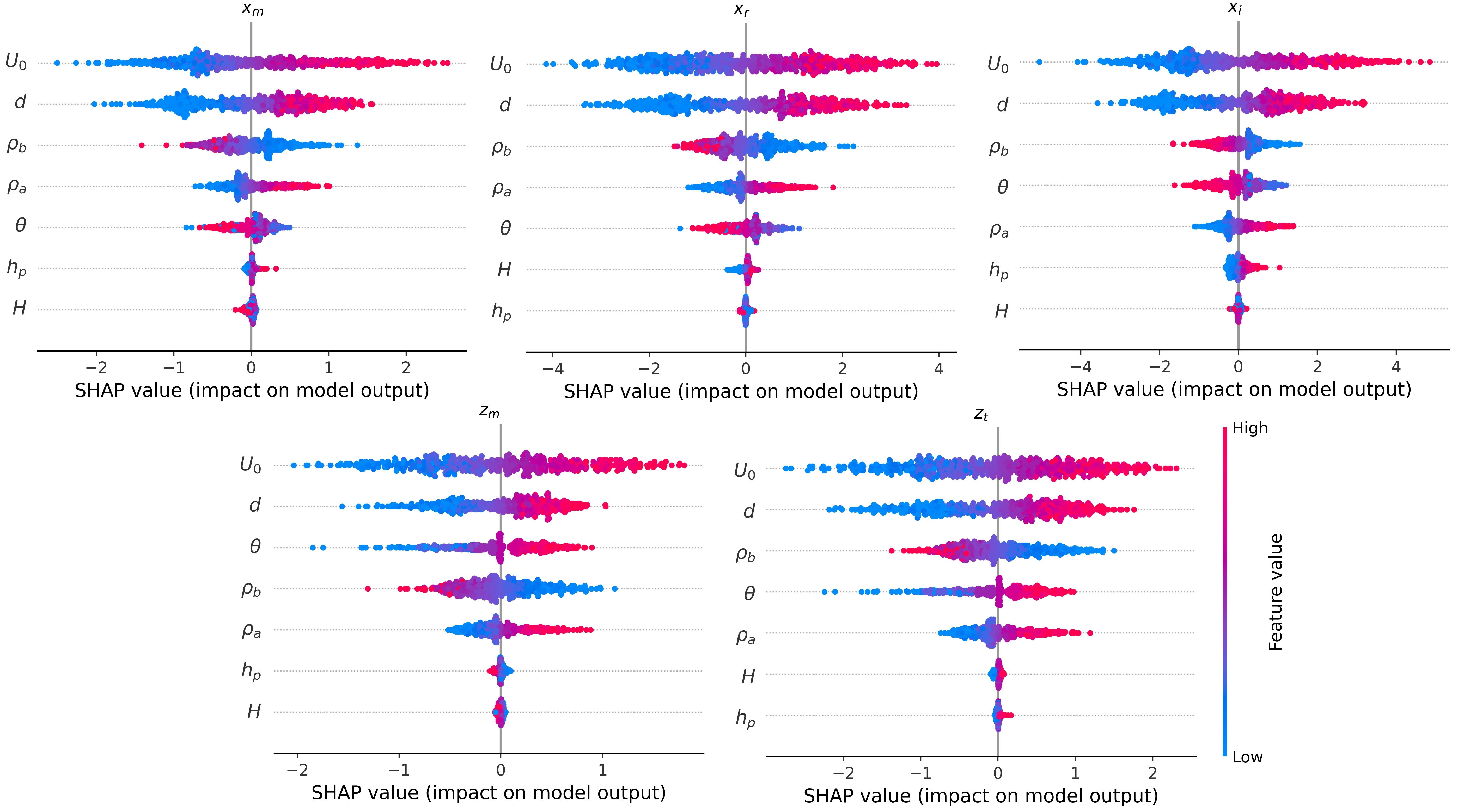}
\caption{SHAP summary plot for a combined dataset on geometrical characteristics.}
\label{fig:shapsum2}
\end{figure}

In Fig. \ref{fig:shapsum2} SHAP summary plot for the combined dataset is presented. The order of importance of the variables is similar to Fig. \ref{fig:shapsummary}. $x_i$ and $z_t$, $\theta$ have smaller importance which can be attributed to more cases with 45$^{\circ}$. Regarding heights, $H$ and $h_p$ have different importance, which is probably due to the dataset with a focus on the shallow water effect or Coanda effect, depending on geometrical characteristics. The importance of the rest input parameters is the same as in the initial dataset. 

\subsection{Reduced ML model}

Most of the previous research analysed geometrical characteristics of negatively buoyant jets with 3 parameters: $\theta$, $Fr$ and $d$. An ANN model could also be trained with the three mentioned input parameters. Results for such model are presented in Table \ref{tab:table_3var}. Predicting technique is the same as in the previous steps. The dataset is built from the same 500 numerical simulations, except input data are compressed. The ANN architecture was tuned for this dataset and the only difference is that it consists of 20 neurons in each of 3 hidden layers, instead of 25 (as seen in Table \ref{tab:hyper}).

It can be seen that the accuracy is similar to that in Table \ref{tab:result2}, therefore it can be concluded that the use of the number of variables is justified and their application to ML models is also possible. Slightly worse results (average R$^2$ decreased from 0.975 to 0.957 and RMSE increased from 0.279 to 0.369) can be attributed to the influence of the remaining variables that are not compressed and the specificity of each case. This is also a possible explanation for some discrepancy in results from previous literature, considering that they mainly observed a reduced number of variables while comparing results. More importantly, this approach does not provide a detailed insight into the importance of each feature using the SHAP method, reducing the explainability of the models essential for understanding the flow behaviour of inclined dense jets.

\begin{table}[!ht]
\caption{Results of ML models for testing (K-fold cross validation) and validation for a reduced number of input variables.}
\centering
\label{tab:table_3var}
\begin{tabular}{ccccc}
        & \multicolumn{2}{c}{R$^2$} & \multicolumn{2}{c}{RMSE} \\ \cline{2-5} 
        & K-fold         & Val.     & K-fold         & Val.    \\ \hline
$x_{m}$ & 0.939$\pm$0.014    & 0.946    & 0.351$\pm$0.057    & 0.310   \\ \hline
$z_{m}$ & 0.952$\pm$0.013    & 0.966    & 0.232$\pm$0.041    & 0.164   \\ \hline
$z_{t}$ & 0.965$\pm$0.009    & 0.968    & 0.276$\pm$0.046    & 0.241   \\ \hline
$x_{r}$ & 0.960$\pm$0.019    & 0.975    & 0.477$\pm$0.084    & 0.413   \\ \hline
$x_{i}$ & 0.930$\pm$0.021    & 0.931    & 0.669$\pm$0.121    & 0.719  
\end{tabular}
\end{table}

\begin{figure}[H]
\centering
\includegraphics[width=\textwidth]{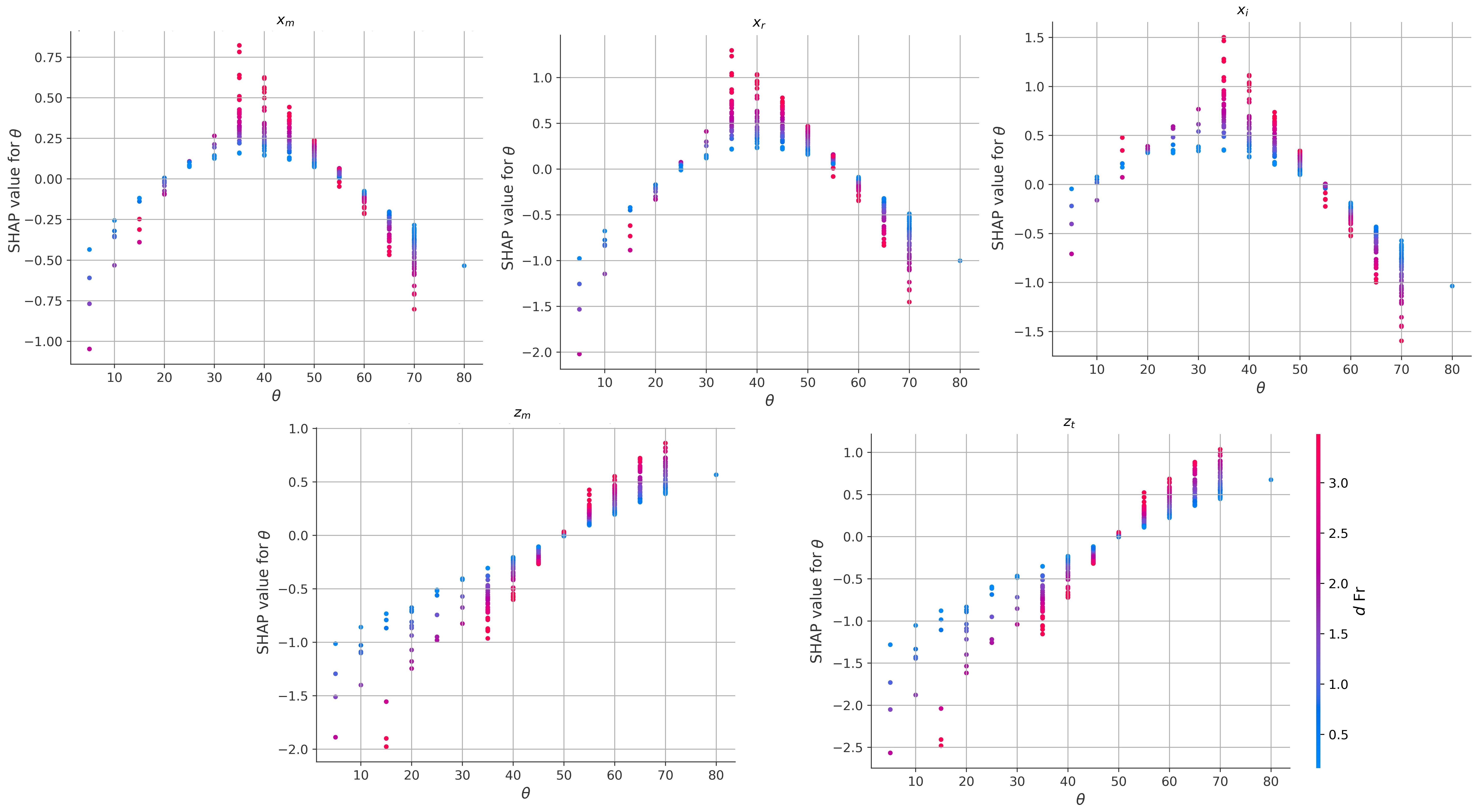}
\caption{SHAP partial dependence plots with a reduced number of variables for $\theta$ and $d$ Fr.}
\label{fig:shapdepp2}
\end{figure}

In Fig. \ref{fig:shapdepp2}, the partial dependence plots are shown for the geometrical characteristics of inclined dense jets. It can be seen that the relationship between $\theta$, SHAP value for $\theta$ and $d$ $Fr$ is similar to previously plotted dependence on $U_0$ in Fig. \ref{fig:shapdepp}. This is consistent with scientific knowledge that the greatest influence on the Froude number is the velocity $U_0$. Therefore, this is a confirmation that it has the most impact on the form and behaviour of inclined dense jets.

\section{Conclusion}

In this paper, a detailed analysis of negatively buoyant jets with interpretable ML algorithms was presented. Numerical simulation provides better prediction than mathematical models compared to experimental data and they allow simulation of different phenomena and flow conditions, but require more time and computational resources. Therefore, in this paper, a dataset of ML models was built based on a large number of CFD simulations with numerical setup validated and compared with previous experimental data with high accuracy. The calculation time of one numerical simulation is long, so complex consolidated databases on which ML models are trained may replace these simulations in the future. The main contributions of this work are:

\begin{itemize}
\item Different types of ML algorithms were tested. The best performance was obtained with ANN for all geometrical points.

\item To get more detailed insight into the performance of negatively inclined jets, the SHAP algorithm was used to analyze the negative or positive contribution of input variables on geometrical characteristics. The most significant parameters were $U_0$, while $H$ and $h_p$ had shown very small importance.

\item By investigating specific cases where Coanda and shallow water effect appear, an increase in the influence of $H$ and $h_p$ on horizontal coordinates are detected.

\item Prediction of ML models based on a dataset with incorporated previous experimental results indicates similar results and behaviour. For some points, the influence of $\theta$ was decreased due to similar angles of experimental data.  

\item Using a reduced number of variables for the prediction of ML models resulted in slightly poorer prediction capabilities, due to neglecting certain effects. 
\end{itemize}

\section*{Acknowledgements}

This research article is a part of the project Computational fluid flow, flooding, and pollution propagation modelling in rivers and coastal marine waters--KLIMOD (grant no. KK.05.1.1.02.0017), and is funded by the Ministry of Environment and Energy of the Republic of Croatia and the European structural and investment funds. 
The authors also acknowledge the support of the Center of Advanced Computing and Modelling (CNRM), the University of Rijeka for providing supercomputing resources for numerical simulations. The work of doctoral student Marta Alvir has been fully funded by the Croatian Science Foundation. 

\bibliographystyle{elsarticle-harv} 
\bibliography{cas-refs}

\end{document}